# Embodied Flight with a Drone

## A. Cherpillod, S. Mintchev, and D. Floreano

*Abstract*—Most human-robot interfaces, such as joysticks and keyboards, require training and constant cognitive effort and provide a limited degree of awareness of the robots' state and its environment. Embodied interactions, instead of interfaces, could bridge the gap between humans and robots, allowing humans to naturally perceive and act through a distal robotic body. Establishing an embodied interaction and mapping human movements and a non-anthropomorphic robot is particularly challenging. In this paper, we describe a natural and immersive embodied interaction that allows users to control and experience drone flight with their own bodies. The setup uses a commercial flight simulator that tracks hand movements and provides haptic and visual feedback. The paper discusses how to integrate the simulator with a real drone, how to map body movement with drone motion, and how the resulting embodied interaction provides a more natural and immersive flight experience to unskilled users with respect to a conventional RC remote controller.

## I. INTRODUCTION

There are several situations where humans are required to control distal robots, for example in exploration of remote areas, inspection, or monitoring of distaster areas. Human–robot interfaces (HRI) can significantly improve the symbiosis between the human and machine [1]. However, most current HRIs—such as joysticks, keyboards, and touch screens—require user training and concentration during operation; therefore, they are limited by a physical and cognitive effort that is often a barrier between the user and the robot. Force feedback through a joystick or through an exoskeleton [2] can improve human robot interaction, but is often limited to direct mappings between human body parts and corresponding body parts of an anthropomorphic robot (a fingered gripper, a robotic arm, a humanoid, e.g.).

Our research goal is to develop novel embodied interactions (instead of "interfaces") with bidirectional feedback between a human and a distal robot with non-anthropomorphic design, sensors, actuators, and behavior. The ultimate goal is not only a more natural and effective control of distal robots, but also the physical transformation of a human into a sensory–motor system (robot) with a different morphology and behavior. With embodied interactions based on a more natural form of control for humans, for example, through gestures, and an immersive and rich multi-modal feedback, the user can ultimately embody a robot with non-anthropomorphic morphology; hence, the user's body and perception can be seamlessly blended with a distal machine. Embodiment, also known as "experiencing presence" in the field of virtual reality [3], results from a combination of self-localization—when the user projects himself onto the distal robot—and self-

ownership—when the user feels the artificial body as his/her own [4], [5]. Embodiment is usually triggered by tailored sensory–motor stimulations that enhance the feelings of self-localization and self-ownership [4], [5]. In proper embodied interactions, the distal robot becomes transparent to the user, whose actions will be defined by only the environment [6]. For example, a user embodying a drone should experience a sensation of flight.

Among the various mobile robots with non-anthropomorphic morphology, drones are interesting candidates for embodied interactions because of their remarkable capability to extend human perception and range of action. Furthermore, remote operation of a drone remains a highly challenging task that can often be mastered by only young video gamers or highly trained people. A truly embodied interaction will not only make the control of drones natural and intuitive even for novices but also provide users with a sensation of flight as if they were there in the air.

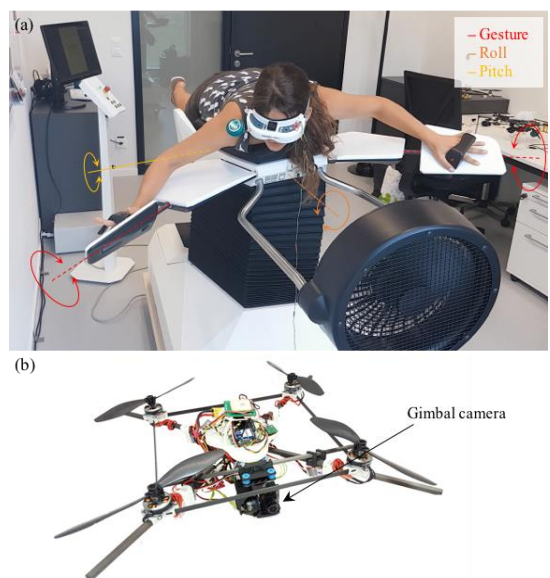

Figure 1. The main components of the setup that allows users to "fly like a drone". The flight simulator (a) has visual, haptic and vestibular (pitch and roll) feedback. The pronation and supination movements of user's hands (red arrows) are transformed into drone motions and the movement of her head is reproduced by the camera gimbal of the drone (b). The pitch and roll angles of the drone are transformed into pitch and roll angles of the flight simulator, its speed is transformed into speed of the air fan, and the image from the drone camera is projected in the visual googles.

In this paper, we demonstrate the advantage of replacing a standard control interface, an RC remote controller (Fig. 3) with a natural and immersive setup for embodied interactions

The authors are with the Laboratory of Intelligent Systems (http://lis.epfl.ch) at Ecole Polytechnique Fédérale de Lausanne (EPFL), CH1015 Lausanne, Switzerland (email: alexandre.cherpillod@epfl.ch).

that allows users to fly with their bodies through an outdoor drone (Fig. 1). To this end, we couple Birdly® (Somniacs[SA], Zurich, Switzerland), a flight simulator that reproduces the flight of a bird with flapping wings in a virtual reality environment, with an outdoor drone. Birdly® allows the user to fly in virtual reality with hand gestures and with movements of the arms that reproduce flapping wings and provides haptic and vestibular feedback on the attitude and speed of the simulated bird. Instead of using a virtual reality simulation of a bird, here we couple the platform to a real outdoor drone. To simplify the experience, we do not require the user to flap the arms to gain speed and we monitor only the hand movements to control the drone. Furthermore, we provide the user with visual input streamed at low latency from the drone. Several challenges are discussed in the article: the identification of a suitable drone, fixed-wing or hovering platform, comparison of two different strategies for mapping gestures to drone motion, and the integration of low latency communication links between the distal drone and Birdly®.

This paper first reviews the state-of-the-art research in human–drone interaction. Then, our approach to integrate Birdly® with a real drone is discussed. Next, the user experience with the different mapping strategies and different input devices, Birdly® vs RC remote controller, are examined through simulations and an outdoor validation of the system is described. Finally, directions for future work are presented in the conclusion section.

## II. RELATED WORK

In human-drone interaction, we find two types of control methods: non-gestural and gestural control. Hereafter, the word gesture is used to indicate both gestures and postures as well as body movements and static poses.

### A. Non-gestural Control

Non-gestural controllers are widely used in drones. In particular, RC remote controller, joysticks, and touch screens are widely used by all major drone manufactures (DJI, Parrot, 3D Robotics). Despite their popularity, these interfaces are not immersive and sometimes difficult to use. For example, it was found that people could not complete their flying tasks using joysticks and touch interfaces (iPhone), whereas they could successfully complete these tasks using upper body gestural control [7]. Touch interfaces were also found to be less enjoyable than upper body gestural control [8]. Brain–machine interaction (BMI) has also been used to control drones. Akce et al. [9] developed a controller based on electroencephalographic (EEG) signals. The user had to think left or right to turn the plane in the simulation. LaFleur et al. [10] developed a similar interface but included the up and down maneuvers control in a real drone. However, the use of EEG signals requires the users to maintain a high degree of concentration during the flight in order to reduce the noise in the EEGs signals. Moreover, the use of EEG signals requires time-consuming user calibration [11]. Gaze gesture control [12] and gaze gesture in combination with EEG signals [11] have also been used in drones. However, this technique requires the user to focus his/her gaze for the control of the drone and prevents the user from exploring the environment.

### B. Gestural Control

Studies on the gestural control of drones can be divided into two main categories: third-person view and first-person view (FPV). In the third-person view, the user acts on the drone as an external viewer, whereas in the first-person view, the user acts from the perspective of the drone.

In the third-person view, the user usually exploits body gestures to interact with single [13] [14] or multiple drones that are within the line of sight [15]. The inherent issue with third-person view is the lack of immersion as the user acts on the drone from an external perspective.

In the first-person view, head motion is commonly used to directly control the drone [16] [17] [18] (Fig. 2(a)). The user usually wears goggles to receive feedback from the video captured by the drone. Nevertheless, head control systems have limitations because they do not allow the user to explore by looking around; the user could lose track of his/her desired trajectory or even crash by just looking at the environment. Another approach is to fly a drone using upper body gestures. Sanna et al. [7] developed a mapping strategy in which a quadcopter was controlled by the gestures of the torso and arms. A constant velocity was added to the quadcopter when the posture of the user approached a threshold. For example, leaning forward more than 15° was firing a forward flight. Different thresholds can be tuned according to different users' sensitivity.

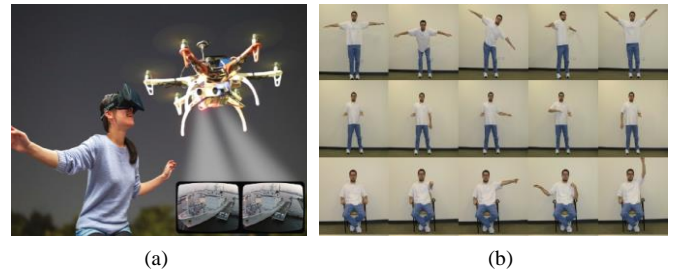

Figure 2. Different types of gestural controls found in the literature. (a) The user controls the position and orientation of the drone with the movement of his head [17]. (b) Different flight styles to control the drone [8].

Pfeil et al. [8] also developed a controller based on torso and arm gestures, but they defined a set of five different gestural controls with their own thresholds (three of them are shown on Fig. 2(b)). One flight style called *first person* was designed as if the user was a bird with arms wide open. Another called *proxy* was designed as if the user was moving the quadcopter in his hands. They performed the task of flying a drone through waypoints to compare the five flight styles. Based on the performance and the appreciation of the user, they concluded that some gestural controllers are more suitable for non-recreational purposes, whereas others are more suitable for recreational purposes. Sakamoto et al. [19] tested an adapted version of the *first person* method developed by Pfeil et al. [8] but in the laying position. The user wore an Oculus Rift to obtain video feedback from the onboard frontal camera of the quadcopter. As soon as the posture approached a given threshold, a constant velocity was added to the quadcopter. However, these gestural controls have two main drawbacks. First, they all use discrete mapping (i.e., every motion of the drone is defined based on the threshold of the gestures). This leads to discontinuous movements of the drone, which significantly diminishes the experience of embodiment because the *immediacy of control* is limited [3], that is, subspaces of the control input have no appropriate consequences. Second, they do not use any vestibular feedback. Therefore, the user could lose the spatial orientation of the drone. By coupling vestibular and visual feedback, the disorientation can be attenuated to favor an immersive and natural flight experience.

Several platforms allow users to simulate flight using gesture commands while receiving both visual and haptic feedbacks. For example, the *Humphrey* developed by Formquadrat [20] is a platform in which the user is hanged in the air with ropes in the laying position facing down with arms wide open. The user flies in the simulation and sees himself/herself from the back as an avatar in a head-mounted display. To control his flight, the user moves his/her arms and at the same time receives a vestibular feedback in roll and pitch owing to movements of the ropes. Another example is the aforementioned Birdly®, where the user flies a simulated bird in a 3D environment. The Humphrey and Birdly® are quite immersive platforms but are only used in simulations. We want to enable their practical use by allowing the user to fly a real drone.

## III. IMPLEMENTATION

At present, Birdly® is one of the most immersive and natural virtual reality flight simulator. Somniacs[SA] developed this platform to allow the user to embody a virtual bird. The user lays on the platform with arms wide open and can control the flight using hand gestures (Fig. 1). The pitch of the bird is controlled by tilting both hands upward or downward. The roll of the bird is controlled by tilting the two hands in opposite directions. Hands can be specifically tilted to combine the pitch and roll commands in order to achieve vertical and lateral movements. The user receives visual, sound, and vestibular feedbacks. The visual feedback is given through Oculus Rift DK2 in a stereoscopic view. The haptic feedback is the airflow coming from the front fan with an air speed proportional to the forward speed. The vestibular feedback is produced by tilting the entire platform in roll and pitch and vertical heave motions. Fig. 1 shows the roll and pitch motions; the heave motion is not integrated in our experiment because there is no corresponding motion with the flight dynamics of our drone. Moreover, the visual feedback with the real drone is produced with Fatshark Dominator HD2 in monoscopic vision for reduced latency (section V).

We want to show that flying using Birdly® is more natural and immersive than flying using a standard RC remote controller (Fig. 3). This interface lacks of vestibular feedback and uses finger gestures, deflection of the right stick, to control pith and roll of the drone. Our approach is to first determine which type of drone, fixed-wing or hovering, is more compatible with the gestures and the vestibular feedback provided by Birdly® and then examine mapping strategies between hands gestures and drone control commands. Finally, the flight performance using Birdly® and the remote controller will be quantified and compared.

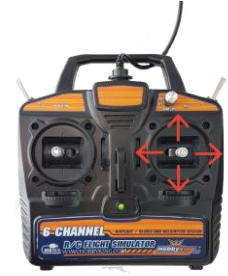

Figure 3. RC remote controller. The two degree of freedom of the right stick allows to give roll (left and right motion) and pitch (up and down motion) commands on the controlled flying platform.

### A. Type of Drone

The shape and flight dynamics of a fixed-wing drone intuitively correspond to the user position and gesture and to the vestibular feedback provided by Birdly®. For example, the eBee® from SenseFly (Cheseaux-sur-Lausanne, Switzerland) is a fixed-wing drone that is steered using two flaps for controlling pitch and roll. Therefore, it is easy to map the hand rotations measured on Birdly® wing paddles (Fig 1a) into fixed-wing flaps movements and use the vestibular feedback of Birdly® (pitch and roll) to replicate the attitude of the drone.

However, hovering drones can be quickly slowed down or stopped in case of an emergency, whereas fixed-wing drones must continue to move and maintain the same speed to stay at the same height. Therefore, we decided to interface Birdly® with a camera-equipped hovering drone that has a modified flight behavior to render, from the perspective of the human user, the dynamics of a fixed-wing drone. To mimic fixed-wing roll and pitch commands the quadcopter, velocity commands are used. Fig. 4 shows qualitatively their relation and Appendix A describes their analytical relation. A key component to give the user the impression to fly a fixed-wing drone is a gimbal system that decouples the camera from the dynamics of the hovering drone. The gimbal is programmed to constantly align the camera with the velocity vector of the drone. For example, in Fig. 4(a), the gimbal aligns the camera with the climbing direction of the drone.

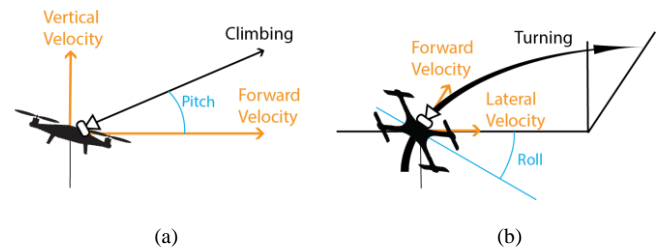

Figure 4. Schematic of a quadcopter mimicking a fixed-wing drone. Velocity commands for a quadcopter can produce attitude commands for a fixed-wing drone. (a) The pitch angle is governed by the forward and vertical velocities. The gimbal camera aligns to the angle of climb. (b) The roll angle is governed by the forward and lateral velocities.

### B. Mapping strategy

The hand rotations on Birdly® wing paddles must be mapped into fixed-wing controls. One possibility is to consider the hands as the flaps of the fixed-wing drone. In this case, a given deflection will translate into an angular acceleration of the fixed-wing drone until reaching a constant angular velocity. Thus, we could associate this mapping with an *angular velocity controller*. This controller is usually called *Acro* mode, e.g., in 3DR products and ArduPilot documentation. It is known to be used by experienced pilots, for example, using remote controllers in drone racing. However, if we identify the user's hand as the frame (or body) of the fixed-wing drone instead of its control surfaces (or flaps), the deflection of the hands would result in a corresponding angle, or *attitude*, of the fixed-wing drone. This control mode is called the *Stabilize* mode in 3DR products and in ArduPilot documentation.

## IV. FROM RC REMOTE TO GESTURAL CONTROL

We compare the use of a standard RC remote controller to the gestural control of the Birdly® given a fixed-wing drone and two different mapping strategies, angular velocity and attitude controller. In order to precisely compare the four combinations, we used a custom-designed virtual simulator of a fixed-wing drone. Fig. 5 shows the simulated environment developed in Unity3D [21]. This comparison aims to determine the most natural and immersive controller by evaluating users' capability to fly through a series of waypoints, which are visualized as small clouds scattered in the sky at different altitudes and directions.

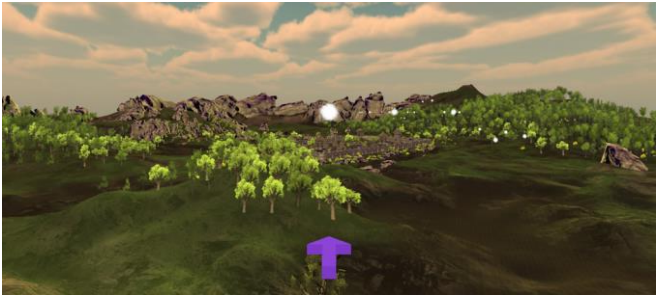

Figure 5. Flight Environment. The user is required to fly through a series of white points visualized as small clouds in the sky. The arrow on the bottom of the screen always shows the direction of the next waypoint, which is helpful if the user turned away from the waypoint and the cloud is not visible. Flight performance is inversely proportional to the average distance from the centers of the clouds.

Human subjects are first explained how the simulated drone is controlled through Birdly® or the RC remote controller, and are then exposed to three successive flight phases: passive flight, training and evaluation. The first phase is a 1-min *passive flight* where the user sees the environment, getting familiar with it, and receives only feedback (visual, haptic, vestibular, and auditory for the Birdly®, visual feedback only for the remote controller). He cannot control the simulated drone. During this first phase, the simulated drone flies autonomously through a sequence of clouds and the subject is explained the task he will have to do in the next phases which is to "follow the clouds and to be as close as possible to the center". The subject then starts a 9-min *training phase*, which involves flying through the clouds one after the other. The subject can ask for breaks, during which the simulation is paused. The training phase is followed by an *evaluation phase*, which involves flying through 84 waypoints; since the drone speed is maintained constant at 12 m/s, the average duration of this phase is approximately 5 min. The spacing between the waypoints is approximately 40 m. The size and flight dynamics of the simulated drone reproduce the fixed-wing drone eBee (senseFly SA). Whenever the subject crashes the drone, he is repositioned in front of the next waypoint.

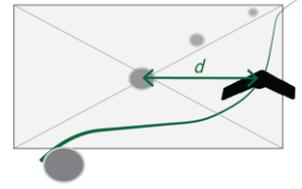

Figure 6. Drone flying through waypoints (grey discs). The distance between the center of the waypoint and the flying platform is measured for each waypoint. This distance lays in the plane perpendicular to the previous and next waypoint (in grey).

We used a quantitative and a qualitative measure to compare the four combinations. The quantitative measure was based on the distance between the drone and the center of each waypoint. To pass a waypoint, the drone had to cross a virtual plane associated with the waypoint (see Fig. 6). 100% performance was obtained when the drone was crossing the center of the waypoint; this value decreases with distance from the center according to a Gaussian function and reached a value of 1% at 38.4 m. This distance has been computed with the data of the evaluation phase and corresponds to the average distance to the waypoints center plus 2.5 times the standard deviation. This is assumed to be an outlier distance [22]. For the qualitative measure, after the experiments subjects were asked to rate a number of statements shown in Table III to indicate their degree of agreement (from 1 to 7). The statements were intended to assess the level of immersion and the naturalness of the control strategy.

### A. Results

42 subjects (37 male, 5 female, age range 19-51 years, average age 28 years) participated in the experiment. 2 subjects felt nauseous (1 male, 1 female) and discontinued the experiment. Among the 40 valid subjects, 10 reported to already have directly or remotely piloted an aircraft for more than one hour. Out of the 40 valid subjects, 20 tested the RC remote controller and 20 the Birdly®. For both interface 10 were presented with the angular velocity controller and the other 10 were presented with the attitude controller, leading to four different combination of interface and mapping strategy. Fig. 7 illustrates the performance of the subjects with the two interfaces and the two mapping strategies during the training phase. For ease of visualization, the curves represent an average window over 20 waypoints. Table I shows the corresponding average performance at the beginning and end of the different trainings.

From Fig. 7, we notice that globally subjects improved their skills all along the training. For each combination, the Mann-Whitney U-test confirmed that there is a significant difference between the starting and ending performance, see Table I. According to Wigdor [23], a natural controller must make the user an expert in a short training period. From the training, we can see that the attitude controller using Birdly® has an average starting and ending performance higher than the other combinations. Indeed, its starting and ending performance are significantly higher than the attitude and angular velocity controller using the RC remote controller and the angular velocity controllers using Birdly® (p < .001).

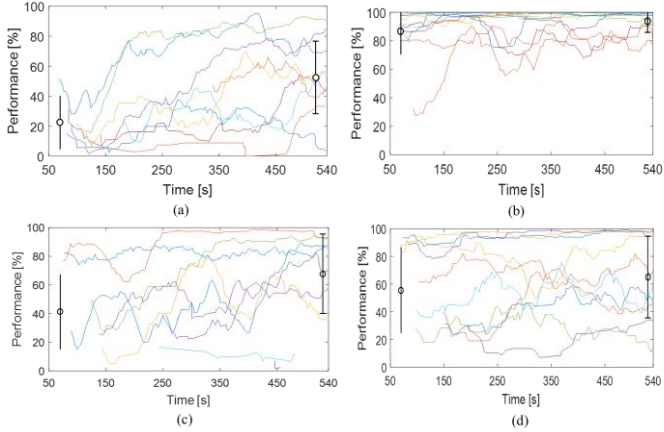

Figure 7. Training performance. Curves represent an average window over 20 waypoints. The mean and standard deviation of the starting and ending performance are shown with error bars. Birdly® with angular velocity controller (a) and with attitude controller (b). RC remote controller with angular velocity controller (c) and with attitude controller (d).

TABLE I. Average starting and ending performance during the training phase. The Mann-Whitney U-test indicates if the starting and ending performances are significantly different.

| Control | Mapping strategy | Average starting performance [%] | Average ending performance [%] | Mann-Whitney U-test |
|---|---|---|---|---|
| Birdly® | Angular velocity | 22.4 | 52.5 | p < .001 |
| | Attitude | 86.9 | 93.5 | p < .001 |
| RC remote controller | Angular velocity | 41.1 | 67.8 | p < .001 |
| | Attitude | 55.6 | 65.0 | p < .001 |

Fig. 8 shows the average performance levels for each subject during the evaluation phase. Table II shows the average performance of the evaluation for each type of interface and mapping strategy. The results show that except the attitude controller using Birdly® the other combination have a higher variability in term of performance and contain subjects with a performance below 50%. The attitude controller using Birdly® presents an average performance which is significantly higher compared to other combination (p < .001). These data from the evaluation confirm that the using Birdly® with an attitude controller is more effective and more natural, according to Wigdor [22] as the level of expertise is higher. Interestingly, the performances of both mapping strategies using the RC remote controller are not significantly different (p > 0.1) in the case of using Birdly® they are significantly different (p < .001). This demonstrates the difficulty encountered by the users using an RC remote controller independently of the mapping strategy.

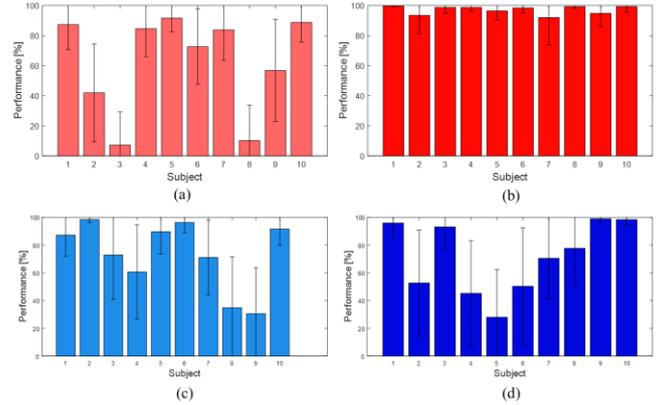

Figure 8. Average performances of each subject during the evaluation phase with their standard deviations. (a) Birdly® with angular velocity controller. (b) Birdly® with attitude controller (c) RC remote controller with angular velocity controller. (d) RC remote controller with attitude controller.

TABLE II. Average performance during the evaluation

| Control | Mapping strategy | Average performance [mean ± SD] |
|---|---|---|
| Birdly | Angular velocity controller | 63.2% ± 38.1 |
| | Attitude controller | 97.3% ± 8.1 |
| RC remote control | Angular velocity controller | 74.7% ± 33.5 |
| | Attitude controller | 71.1% ± 37.0 |

The results of the questionnaire are presented in Fig. 9. The relative questions are shown in Table III. From the training and the evaluation we saw that the angular velocity controller using the RC remote controller has in both cases the second best performance, see Table I and Table II. Now regarding the questionnaire, people found that using Birdly® with the attitude controller was giving a better sensation of controlling the flight trajectory (question index #2) (p < .05) and they enjoyed it more (question index #6) (p < .05). This confirmed that the attitude controller using Birdly® is more immersive and natural. Indeed according to Wigdor [23], the enjoyability of the interface is another factor of its naturalness.

TABLE III: Questionnaire filled out by the subjects at the end of the experiment. Subjects provide ratings between 1 (Disagree) and 7 (Strongly agree).

| Index | Question |
|---|---|
| 1 | I felt as if I was flying myself. |
| 2 | I had the sensation of controlling the flight trajectory. |
| 3 | I felt some physical discomfort. |
| 4 | If I could fly in the air, I would use the same gesture. |
| 5 | The proposed gesture control was natural for me. |
| 6 | I enjoyed the experiment. |
| 7 | I found the experiment tiring. |

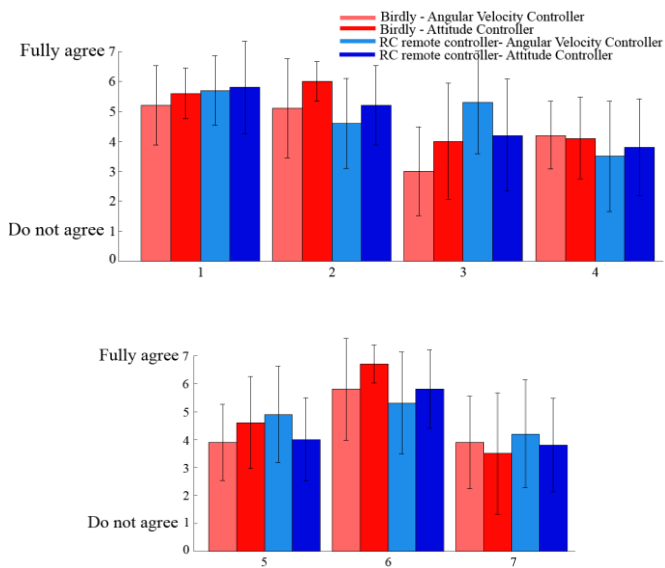

Figure 9. Questionnaire results. Mean and standard deviation. The questions are presented in Table I. Higher agreement is better, except for question 7.

## V. VALIDATION WITH A DRONE

Based on those results, we validated the attitude controller by integrating Birdly® with a real outdoor drone (Fig. 10). Birdly® sends the angle of the user's hands to a laptop. The laptop converts these angles into velocities of a fixed-wing drone, as described in Appendix A, and sends these velocities to the quadcopter. The laptop also receives the user's head orientation from an inertial measurement unit and sends it to the quadcopter. The quadcopter flies according to the received velocity commands and sends the head orientation commands to the gimbal camera with correct angles to face the direction of motion as if it was mounted on a fixed-wing drone.

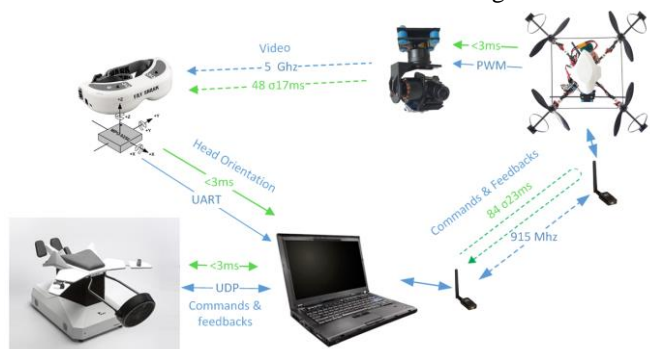

Figure 10. Schematic of the interface. The connections are shown in blue with their respective communication support. The delays are shown in green.

The drone camera, a Sony 700TVL CCD, streams the video to Fatshark Dominator HD2 goggles. The quadcopter in the meanwhile sends its measured attitude and velocities to the laptop. The laptop transforms these values to mimic the behavior of a fixed-wing drone and sends them to Birdly®. The quadcopter autopilot (MAVRIC[1]) handles all the low-level commands. The gimbal camera possesses two degrees of freedom, which are pitch and yaw. This allows the user to look up and down and right and left with respect to the current direction of forward flight.

### A. Latency

To improve the immersion, care has been taken to reduce communication latency for the commands, feedback, and video. For the commands and feedback communication, several wireless devices have been tested and their latency has been measured. Fig. 11 shows the resulting latencies as a function of the frequency of emission. The measured latency is the time between the emission of commands from the laptop and the reception of them through the feedbacks using the Mavlink protocol [24].

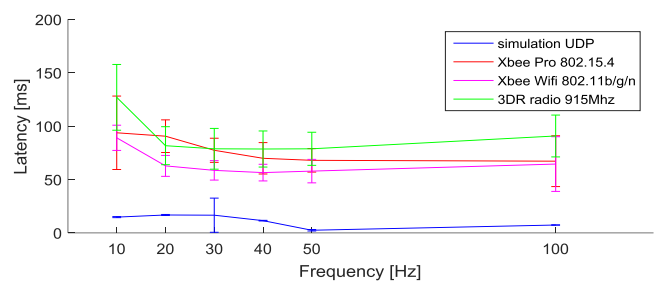

Figure 11. Communication latency for multiple wireless devices. Time between the emission of the commands from the laptop to the drone (21 bytes) and their reception through the feedbacks (56 bytes). The commands and feedbacks are emitted at the same frequency.

These experimental data show that Xbee Wifi has the smallest latency of 56.5 ms. For frequencies greater than 20–30 Hz, the latency does not change but the number of lost packets increases. According to [25], the latency of the commands and feedbacks (haptic feedback) must be below 30–50 ms to perceive feedback as instantaneous. A higher latency would reduce the immersion as the *immediacy of control* would not be respected [3]. Ideally, the frequency should be infinite in order to prevent additional delays between two samplings. We chose a frequency of 30 Hz as the highest frequency with the smallest number of packet losses on the communication link. Table II presents the robustness of the wireless devices. The Xbee solutions are less robust in case of obstacles in the line of sight. This was qualitatively evaluated considering the high number of lost signals the Xbee solutions had during the flights. With its lower frequency of 915 MHz, the 3DR radio module allows transmission despite the obstacles in the environment. Therefore, we chose the 3DR radio module for connecting the drone to the laptop.

TABLE II: Percentage of lost packets between the emission and the reception at a distance of 100 m for different solutions. Packets are sent at 30 Hz from both the laptop and the drone.

| Wireless Device | Packets loss [%] @100m line of sight | Packet loss @100m with obstacles |
|---|---|---|
| Xbee Pro 802.15.4 | 2.6 | high |
| Xbee Wifi 802.11 b/g/n | Out of range | Out of range |
| 3DR radio 915Mhz | 2.0 | low |

---

[1] https://github.com/lis-epfl/MAVRIC_Library

Several video streaming solutions for the video feedback are listed in Table III with their relative latencies. According to [26], the visual feedback should be given in less than 50 ms to ensure natural immersion. Three solutions would comply with this criterion. However, since the CONNEX ProSight HD was not released at the time of the experiment and the DJI Lightbridge2 needs a DJI remote controller to receive the video signal, these two solutions were not considered. Therefore, we chose the FPV camera with the Fatshark Dominator goggle for video display. A latency of 48 ms was obtained. The reproduction of the head motion by the gimbal camera should be delayed by less than 50 ms as well; in our case, however, the gimbal camera system had a latency of around 100ms. This latency is mainly due to the latency of the 3DR radio module between the laptop and the quadcopter and the latency of the visual feedback. The consequence is a mismatch between user head motion and visual feedback reducing the immersion as the immediacy of control is respected [3].

TABLE III: Latency and quality of various video streaming solutions commonly used in FPV. *From datasheet.

| Video Streaming Solution | Quality | Average latency [ms] |
|---|---|---|
| Parrot Bebop I w. Computer FHD | 1080p | 162 |
| Parrot Bebop I w. Iphone 6 FHD | 1080p | 258 |
| DJI Phantom 3 w. Iphone 6 HD | 720p | 263 |
| 3DR Solo w. proprietary viewer | 720p | 180 |
| Sky Drone w. Oculus Rift FHD | 1080p | 150 |
| CONNEX ProSight HD | 720p | 26* |
| FPV camera w. Fatshark goggles ~VGA | 600TVL (between 480p&720p) | 48 |
| DJI lightbridge2 HD | 720p | 50* |

### B. Flight Validation

We validated the system presented in Fig. 10 with an outdoor flight. We installed Birdly® inside a building facing the flight area on EPFL campus (Fig. 12). The subject was asked to continuously fly along an 8-shape trajectory around two vertical obstacles made of red helium balloons at 10 m from the ground. For safety reasons, if the subject attempts to fly below 20 meters of altitude, a repulsive force is added to the vertical velocity of the drone and can be sensed by the subject as a body movement correction. The flight lasted 5 min. The subject was a 22-year-old male without previous experience in drone control. After the flight, the subject answered the questionnaire used for the comparison of gesture controllers (see Table I).

As shown in Fig. 12, the naïve subject could fly rather precisely and repeatedly around the two balloons. He maintained an altitude between 17 m and 20 m. This happened because he was constantly trying to fly downwards but was automatically repulsed upwards. The subject reported that he was not conscious of attempting to fly towards the ground. A possible explanation is that the two obstacles were below 20 m and the subject wanted to precisely fly around them. Another possible explanation is that at lower altitude, the optic flow is stronger, which gives more accurate information about the drone's speed, allowing better control decision.

In the questionnaire (Table I), the subject strongly agreed that the flight with Birdly® integrated with the real drone was enjoyable (7/7). He agreed that he experience the sensation of controlling the flight trajectory (6/7) and the sensation of flight (5/7). This outdoor flight revealed that the subject could embody the drone by self-identifying and self-localizing himself in the air.

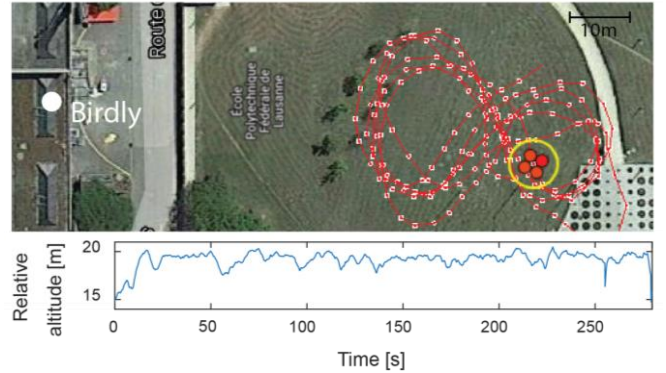

Figure 12. Trajectory of the validation flight. The subject flew several times around two red helium-filled balloons (not visible in the figure). The circled item corresponds to the quadcopter. The distance between Birdly® and the center of the trajectory is 55 m.

### VI. CONCLUSION

In this paper, we proposed the integration of the Birdly® platform with a real drone to enable an embodied interaction that allows people to fly like a drone. We found that using a fixed-wing drone and a mapping between user gestures and drones commands based on an attitude controller enable an immersive and natural flight experience. We also found that using a gestural control based on Birdly® is more natural and immersive than using standard RC remote controller. Moreover, we addressed the technical challenges of interfacing Birdly® with a real drone.

One future challenge is to develop a system more portable than the Birdly® platform but having similar gestural controls and feedback. A promising avenue would be to use exoskeletons for gesture recognition and haptic feedback. Another challenge is to add a shared control that adaptively shift the control between the user and the autopilot of the drone based on the skills or emotional fatigue of the user.

### VII. APPENDIX A

To mimic the flight style of a fixed-wing, velocity commands $v_x, v_y, v_z$ are given to the quadcopter in the semi-body reference system as function of desired roll and pitch angles. The semi-body frame is equal to the earth reference frame but rotating according to the yaw of the platform. Once the constant speed $v$ of the fixed-wing and its desired roll $\phi_{ref}$ and pitch angles $\theta_{ref}$ are chosen, the command value $v_z$ for the quadcopter is computed as:

$$v_z = \tan(\theta_{ref}) \cdot v \qquad (1)$$

Then we compute the value of $v_y$. Its analytical expression is less trivial. According to Fig. 13, we can establish the following relations:

$$\dot{\alpha} = \frac{v}{R} \qquad (2)$$

$$\tan(\phi_{ref}) = \frac{F_c}{F_v} = \frac{m \cdot a_c}{m \cdot g} = \frac{v^2}{R \cdot g} \qquad (3)$$

where $a_c, F_c$ and $F_v$ are the centripetal acceleration, the centripetal force and the counter-gravity force.

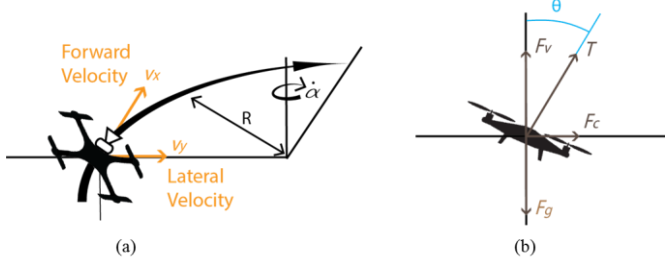

Figure 13. Schematic of a quadcopter mimicking a fixed-wing drone. (a) A lateral velocity induces a roll angle and a radius of curvature. (b) A centripetal force Fc is present when turning.

When a turn maneuver is initiated with a lateral velocity $v_y$, an internal controller acts on the yaw rate $\dot{\psi}$ to make the drone follow a circular trajectory

$$\dot{\psi} = k \cdot tan^{-1}\left(\frac{v_y}{v_x}\right) = \dot{\alpha} \qquad (4)$$

where $k{=}0.6$ is a gain tuned to mimic the dynamic of the senseFly Swinglet. By substituting $R$ from equation (2) into equation (3) and replacing $\dot{\alpha}$ by its expression in (4) we obtain:

$$v_y = v_x \cdot \tan(\frac{\tan(\phi_{ref}) \cdot g}{v \cdot k}) \qquad (5)$$

In order to keep the desired speed $v$ of the fixed-wing constant, the following expression needs to be respected:

$$v^2 = v_x{}^2 + v_y{}^2 + v_z{}^2 \qquad (6)$$

This leads to having a command value $v_y$ as:

$$v_y = \frac{\sqrt{v^2 - v_z{}^2} \cdot A}{\sqrt{1 + A}}, \quad \text{with} \quad A = \tan(\frac{\tan(\phi_{ref}) \cdot g}{v \cdot k}) \qquad (7)$$

And then the value of $v_x$ is obtained with:

$$v_x = \sqrt{v^2 - v_y{}^2 - v_z{}^2} \qquad (8)$$

In order to always have real values for the velocity commands, the value of $v_y$ and $v_z$ are upper bounded right after their computations in equation (1) and (7).

## Acknowledgment

The authors thank the Swiss NCCR Robotics Symbiotic Drone Project team for their advice on the experiment protocol. We also thank the MAVRIC team for the software development of the drone and Grégoire Heitz for his useful advice and time as a safety pilot. This work was supported by the Swiss National Science Foundation through the National Centre of Competence in Research Robotics.